\documentclass[10pt,twocolumn,letterpaper]{article}

\usepackage{ijcb}
\usepackage{times}
\usepackage{graphicx}
\usepackage{amsmath}
\usepackage{amssymb}
\usepackage[norule,symbol,perpage]{footmisc}
\usepackage{multirow}
\usepackage{diagbox}
\usepackage[inline,shortlabels]{enumitem}
\usepackage{tabto}
\usepackage[table]{xcolor}
\usepackage{caption}
\usepackage{subcaption}
\usepackage{arydshln}
\usepackage[export]{adjustbox}
\usepackage{tabularx}

\DeclareMathOperator*{\argmin}{arg\,min}

\ijcbfinalcopy 

\ifijcbfinal\fi

\makeatletter
\def\ps@IEEEtitlepagestyle{
\def\@oddfoot{\mycopyrightnotice}
\def\@evenfoot{}
}
\def\mycopyrightnotice{
{\hfill \footnotesize 978-1-7281-9186-7/20/\$31.00 \copyright 2020 IEEE\hfill}
}
\makeatother

\begin{document}

\title{Is Face Recognition Safe from Realizable Attacks?}


\author{Sanjay Saha\\
Department of Computer Science\\
National University of Singapore\\
{\tt\small sanjaysaha@comp.nus.edu.sg}
\and
Terence Sim\\
Department of Computer Science\\
National University of Singapore\\
{\tt\small tsim@comp.nus.edu.sg}
}
\maketitle
\thispagestyle{empty}

\begin{abstract}
Face recognition is a popular form of biometric authentication and due to its widespread use, attacks have become more common as well. Recent studies show that Face Recognition Systems are vulnerable to attacks and can lead to erroneous identification of faces. Interestingly, most of these attacks are white-box, or they are manipulating facial images in ways that are not physically realizable. In this paper, we propose an attack scheme where the attacker can generate realistic synthesized face images with subtle perturbations and physically realize that onto his face to attack black-box face recognition systems. Comprehensive experiments and analyses show that subtle perturbations realized on attackers face can create successful attacks on state-of-the-art face recognition systems in black-box settings. Our study exposes the underlying vulnerability posed by the Face Recognition Systems against realizable black-box attacks.
\end{abstract}

\let\thefootnote\relax\footnotetext{\mycopyrightnotice}

\section{Introduction} \label{goto:intro}

Face recognition is one of the most convenient and popular biometric authentication systems. Being contactless, applicable for a large crowd at a time, and more importantly, recent improvements in the accuracy of these systems are the main reasons behind the widespread uses of Face Recognition Systems (FRS). Different kinds of attacks on FRS and their countermeasures are frequently studied in the field of biometrics.

Attacks on FRS can be broadly classified mainly into two groups:
\begin{enumerate*}
  \item Presentation attacks.
  \item Adversarial attacks.
\end{enumerate*}
In presentation attacks, the aim of the Attacker is to fool the FRS by presenting a photo of a face or put on a mask. Presentation attacks and detection of these attacks have been very common recently. Many studies\cite{nikisins2018effectiveness, raghavendra2015presentation, ramachandra2017presentation, scherhag2017vulnerability} have been done to develop presentation attacks, and detection techniques to prevent these attacks. Adversarial attacks, on the other hand, rely on generating adversarial face images, or adversarial patches (mostly through Generative Adversarial Networks, Auto-encoders, etc.). Adversarial attacks also extend to other image classification problems. They rely on making subtle changes to the input images (mostly in the forms of noise, small changes in the pixels which are unnoticeable to human eyes). Studies\cite{dong2019efficient, sharif2017adversarial} show that even the state-of-the-art FRS are vulnerable to adversarial attacks. All these studies are helpful for finding out inherent vulnerabilities in FRS. 

We can also group attacks on image classification systems based on how much the Attacker knows about the target system: 
\begin{enumerate*}
  \item White-box attack.
  \item Black-box attack.
\end{enumerate*}
In a white-box attack, the Attacker knows the underlying working principle of the target system. The Attacker can use this knowledge of the internal parameters to modify the decisions of the system. Although these attacks are more successful against targeted systems, it is not realistic to know their parameters. A more practical kind of attack is a black-box attack where the internal parameters are unknown to the Attacker. Hence, the Attacker would need to rely only on the output of the system. 

To better describe our Attack Scheme, let us assume a `Guarded Attack Scenario' where a FRS is used to secure a building. It authenticates users of the building. The system is also attended by a security guard. The Attacker wants to get into the building by fooling the FRS. He also must not raise any suspicion of the security guard (e.g. present a photograph/video of face, put on mask etc.). Hence the attack needs to be realizable. Additionally, he does not know the internal structure of the FRS which means it must be a black-box attack.

It is not practical to launch a presentation attack by showing a printed photo, video, or, put on a mask in front of the security guard as he would immediately notice the attack. In this scenario, it is also not possible to launch an adversarial attack because adversarial images with subtle pixel manipulations cannot be recreated physically. But the attacker can attempt to grow a beard, put on a scar, or make a facial expression to fool the FRS. We call such attacks `physically realizable'. Also, the security guard does not notice because the attack does not raise any suspicion. In this paper, our goal is to address the `Guarded Attack Scenario' by making subtle perturbations on the Attacker's face with the help of a face synthesizer so that the same perturbations can be physically realized. 

In this paper, we explore three types of attacks on FRS in the Guarded Attack Scenario:
\begin{enumerate*}
    \item Break-in.
    \item Impersonation.
    \item Evasion.
\end{enumerate*}
Details on these attacks are given in Section \ref{goto:attack-type}. The contribution of this paper is to show that FRS are vulnerable to realizable attacks, and thus to urge more research in this area.

\section{Related Works}

\textbf{Attacks on objects and face recognition.} Attacks on object classification systems, and FRS have been studied broadly in recent times especially after the advent of deep neural networks. 
Presentation attacks is the most common attack against FRS. Studies \cite{dhamecha2014recognizing, nikisins2018effectiveness, raghavendra2015presentation, raghavendra2017vulnerability, ramachandra2017presentation, scherhag2017vulnerability} on presentation attacks, and detection of these attacks show that many FRS are vulnerable to these attacks. However, these presentation attacks\cite{raghavendra2017vulnerability, scherhag2017vulnerability} are developed for the scenario when the FRS are unattended i.e. there is no security guard near the system to monitor. Thus, these attacks are not useful in an attended scenario like the Guarded Attack Scenario where it is necessary for the attacker to be stealthy. Also traditional presentations attacks (e.g. presenting a photograph, or video) are also not useful in the Guarded Attack Scenario.
Other than presentation attacks, Adversarial studies\cite{brown2017adversarial, goodfellow2014explaining, kurakin2016adversarial, szegedy2013intriguing} show that careful perturbation in images based on pixel manipulations can also fool the classifier to misclassify. Deep Convolutional Neural Network models posses vulnerability against carefully crafted white-box attacks. These\cite{sharif2016accessorize, sharif2017adversarial} attacks are based on manipulation on the face of an attacker using a printed out adversarial eyeglasses. Although successful, these attacks utilized prior information about the models that were attacked. In this study, we are focusing on the black-box scenario where the attacker would not have any information from the model other than its predicted identity, and its prediction score.

\textbf{Black-box attacks.} These black-box attacks\cite{athalye2017synthesizing, brendel2017decision, chen2017zoo, ilyas2018black, papernot2017practical} on object classification models are based on adding subtle perturbations in the input images. However, these attacks are not possible to be replicated in the real world as they are based on pixel manipulations on the images rather than perturbations on the actual objects. In \cite{brown2017adversarial} the authors designed a black-box attack with a physical patch attached near an object which causes wrong classification from the classifier. However, this attack is not subtle and also does not scale to FRS. In \cite{mirjalili2018gender, mirjalili2017soft}, attacks targeting face images have been made using semi-adversarial networks to preserve gender privacy. In \cite{dong2019efficient, goswami2018unravelling}, the identity of the face was the target in black-box settings for attacks on FRS. Although these attacks are making small perturbations on face images, they are pixel-based perturbations that cannot be realized on physical faces.

\textbf{Realizable attacks.}
The closest work to our attack is \cite{sharif2016accessorize}. In their paper, the authors attacked, and succeeded to fool a FRS by putting on a printed adversarial spectacle frame in a white-box setting. Although this attack is realizable, and stealthy, it is only possible in a white-box scenario. On the other hand, perturbations proposed in \cite{dong2019efficient, goswami2018unravelling, mirjalili2018semi} are subtle but are only possible for face images i.e. they cannot be realized on physical faces. Unlike these attacks, in our Attack Scheme, the Attacker attacks a black-box FRS by adding on facial features like beard, marks, makeup, etc.

In this paper, we focus on the Stealthy Attack Scenario as described in Section \ref{goto:intro}. We propose realizable perturbations so that it is possible to physically reproduce them on real faces. Our Attack Scheme succeeded in attacking multiple FRS in the above mentioned scenario with realistic synthesized faces.

\section{Method}

\subsection{The Attack Scheme}
The basic structure of our Attack Scheme has the following major components: a target FRS, a \textit{gallery} of authorized subjects, and a face synthesizer. We present a synthesized image of the Attacker's face to the target FRS which authenticates the face and returns the best-matched identity ($id$) and the corresponding score ($s$). The score can be the \textit{distance} (euclidean, cosine, etc.) of the best-match or, \textit{confidence} in the prediction. For simplicity, let us assume that score, $s, \; 0 \leq s \leq 1$ is the distance with the best-matched face in the \textit{gallery}. Hence, lower $s$ means the FRS is highly confident of its prediction and vice verse. When the score ($s$) is below a predefined threshold $\theta$ the FRS authenticates the Attacker and the attack is considered successful. 

\begin{figure}[h!]
    \centering
    \includegraphics[width=.48\textwidth]{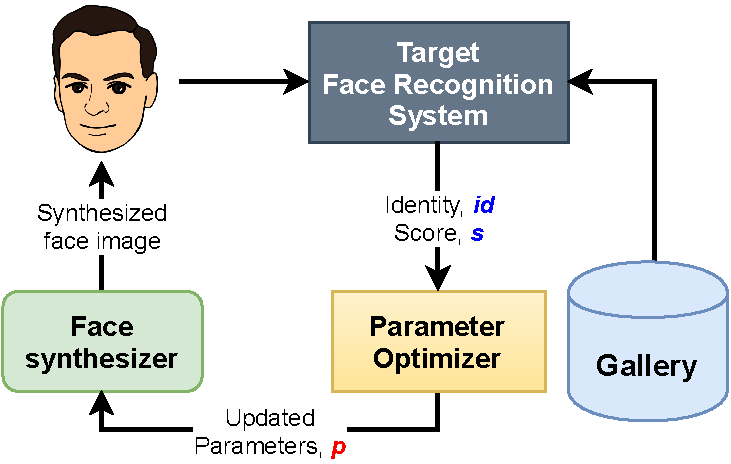}
    \caption{Attack Scheme: synthesis of optimized realizable face and using it to attack the target black-box FRS. The FRS matches the presented face with the face images in the gallery, and returns identity and score.}
    \label{fig:attack-scheme}
\end{figure}

We generate a new face image of the Attacker using the synthesizer by manipulating the controllable parameters, $p$. More details on the face synthesizer, and the parameter vector $p$ are in section \ref{goto:facesynthesizermmda}. The Attack Scheme is an optimization problem where we try to minimize the score, $s$ until $s < \theta$. Hence, we are essentially using an `analysis-by-synthesis' method (Figure \ref{fig:attack-scheme}). 

Let $f: X \rightarrow Y$ denote the target FRS where $X$ is the set of input faces and $Y$ is the set of scores of the best-matched faces. Each element in $Y$ is composed of two parts: $id$ (predicted identity) and $s$ (score). $\mathcal{S}(p)$ be the face synthesizer that generates a new face for the Attacker using the synthesis parameter vector, $p$. Let $\mathbb{P}$ be the parameter vector space. Our goal is to minimize the score ($s$) returned by our objective function $f$. If $p_{min}$ is the optimized parameter vector at the end of an attack on the FRS, and $s_{min}$ is the score for the face synthesized using $p_{min}$, we can define our Attack Scheme as the optimization problem:


\begin{align} \label{eqn:optimization}
p_{min} &= \argmin_{p} (f(\mathcal{S}(\mathbb{P}_p))) \\
&subject ~ to ~ B(\mathbb{P}_p) \leq 0, ~ where ~ B() ~ constrains\notag\\
&the ~  parameter ~ vector ~ \mathbb{P}_p ~ (see ~ Section ~ \ref{goto:newface}) \notag\\
s_{min} &= f(\mathcal{S}(p_{min})) \\
result &= 
\begin{cases}
    success, & \text{if } s_{min} \leq \theta\\
    failed, & \text{otherwise}
\end{cases}
\end{align}

In our experiments, this optimization was done using the Nelder-Mead method in the Scipy optimization package. We initialized the optimization parameter $p_0$ depending on the types of attacks (see Section \ref{goto:attack-type}).

\subsection{Target Face recognition systems (FRS)} \label{goto:frs}
We have used three FRS as targets which are publicly available.
\begin{enumerate*}
    \item Python \textit{face\_recognition}\cite{ageitgey} library.
    \item RESNET50\cite{he2016deep} trained with Oxford's VGGFace\cite{parkhi2015deep}.
    \item Face++\cite{facepp} API.
\end{enumerate*}
Although the system architectures of the first two systems are at public disposal, we treat them as black-boxes. We use the best-matched identity ($id$), and score ($s$) returned by the target systems to optimize our objective function $f$. Details on the thresholds ($\theta$) for different FRS are given in section \ref{goto:setup}.

\subsection{Gallery} 
Target FRS have a \textit{gallery} of faces that they recognize. We have taken a subset of the full-frontal face images from the CMU Multi-PIE dataset\cite{gross2010multi}. The Multi-PIE dataset has 338 people's faces from which we have used a subset of the 338 faces for our experiments. We have made sure that people from different races, skin colors, and genders are there in the \textit{gallery} to ensure that face recognition is not biased to any of these factors.

\subsection{Types of attacks} \label{goto:attack-type}
We carried out three kinds of attack as listed below. All the attacks work in similar way as is described in the Attack Scheme (Figure \ref{fig:attack-scheme}). We refer back to the `Guarded Attack Scenario' from Section \ref{goto:intro} to describe these attack types. A summary of these attacks in terms of variations in \textit{gallery} and \textit{Victim(s)} are presented in Table \ref{tab:attack-types}. 

\textbf{Break-in.}
The Attacker tries to gain access to the secured building in this attack. He is not registered in the \textit{gallery}. He presents his perturbed face to the FRS to fool the system to identify him as one of the faces in the \textit{gallery}. There is no specific Victim in the \textit{gallery}. A break-in attack is deemed successful when the FRS authenticates the Attacker as one of the registered faces in its \textit{gallery}. 

\textbf{Impersonation.}
This attack is similar to the break-in attack as the Attacker is not registered in the \textit{gallery}. However, now the Attacker tries to impersonate someone from the \textit{gallery}. We call the selected person `Victim'. The Attacker tries to make the FRS think his face as the Victim's face. An impersonation attack is successful only when the FRS identifies the Attacker as the Victim and no one else. Even if the Attacker is authenticated as some other person, it is not enough to be considered as successful. Hence, impersonation attacks are more challenging than break-in attacks.

\textbf{Evasion.}
Unlike the previous two attacks, the Attacker is now registered in the \textit{gallery}. The goal of the Attacker is to avoid being identified as himself by the FRS. There are two scenarios:
\vspace{-2mm}
\begin{itemize} \label{goto:evasions}
    \itemsep-0.25em
    \item \textbf{Partial Evasion}: The Attacker knows the people in the \textit{gallery} and wants to impersonate someone else (a Victim). This attack is considered successful when the Attacker is identified as the Victim and no one else.
    
    \item \textbf{Full Evasion}: 
    This is where the Attacker does not want to be recognized as \textit{anyone} in the gallery. This is also the `Police Watchlist Scenario', in which the police use a FRS to identify a watchlist of criminals. The Attacker, who is in the watchlist, needs to avoid being recognized as himself, as well as \textit{anyone} else in the watchlist.
\end{itemize}
\vspace{-3mm}

\begin{table}[h!]
\centering
\resizebox{0.5\textwidth}{!}{%
\begin{tabular}{|l|l|l|}
\hline
\textbf{Attack} & \textbf{Gallery} & \textbf{Victim(s)} \\ \hline \hline
Break-in & \begin{tabular}[c]{@{}l@{}}Attacker is NOT\\ IN the gallery\end{tabular} & No specific Victim \\ \hline
Impersonation & \begin{tabular}[c]{@{}l@{}}Attacker is NOT\\ IN the gallery\end{tabular} & One, selected by Attacker \\ \hline
Evasion & \begin{tabular}[c]{@{}l@{}}Attacker is IN \\ the gallery\end{tabular} & \begin{tabular}[c]{@{}l@{}}No specific Victim as long\\ as it is not the Attacker\end{tabular} \\ \hline
\end{tabular}%
}
\caption{Attack types' summary.}
\label{tab:attack-types}
\end{table}
\vspace{-2mm}

\subsection{Face Synthesizer} \label{goto:facesynthesizermmda}
One of our main focuses in this work is to generate realizable face images so that it can be replicated in real-world scenarios. To get such face images we have used the  Multimodal Discriminant Analysis (MMDA) face synthesis model from \cite{sim2015controllable}. This method of face synthesizing is capable of generating realistic-looking faces from the given different variations of training face images. Other synthesis methods\cite{li2019learning, choi2018stargan} could be used that are based on Generative Adversarial Networks (GANs). But this is not the focus of this work. We use MMDA because it is easy to train and it produces good quality realistic images.

\begin{figure}[h!]
\centering
\begin{subfigure}{.32\linewidth}
  \centering
  \includegraphics[width=.98\linewidth]{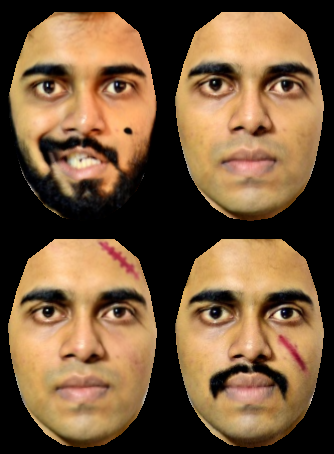}
  \caption{Attacker 1: South Asian Male}
  \label{fig:attact-break-in}
\end{subfigure}
\begin{subfigure}{.32\linewidth}
  \centering
  \includegraphics[width=.98\linewidth]{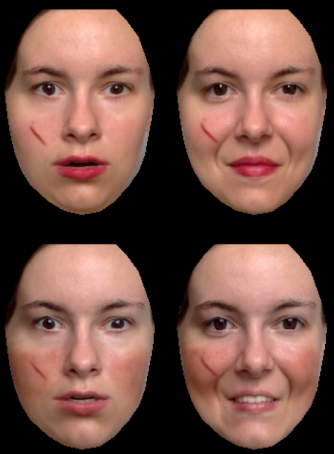}
  \caption{Attacker 2: Caucasian Female}
  \label{fig:attack-impersonation}
\end{subfigure}
\begin{subfigure}{.32\linewidth}
  \centering
  \includegraphics[width=.98\linewidth]{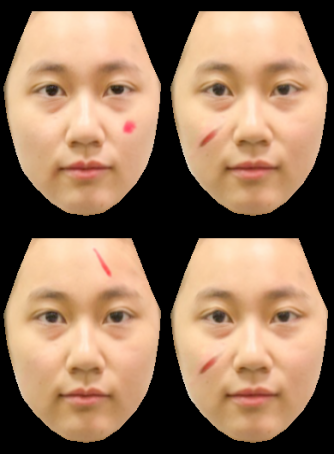}
  \caption{Attacker 3: East Asian Female}
  \label{fig:attack-impersonation}
\end{subfigure}
\caption{Sample images of the Attackers used to train the face synthesizer. They show different combinations of scars, mole, facial hair, expression, etc. The three attackers were chosen from different ethnicity and genders. This helps to generalize the attack scheme in terms of different ethnicity and genders.}
\label{fig:attackers}
\end{figure}

\def\rot#1{\rotatebox{90}{#1}}
\begin{figure*}
    \centering
    \begin{tabular*}{\textwidth}{cccccc}
        \rot{\rlap{~~~~~Attackers}} & \includegraphics[width=30mm]{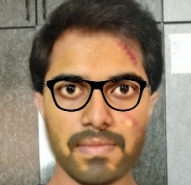} & \includegraphics[width=30mm]{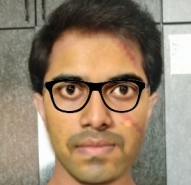} &
        \includegraphics[width=30mm]{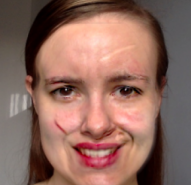} & \includegraphics[width=30mm]{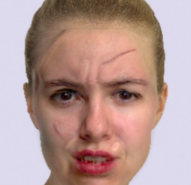} & \includegraphics[width=30mm]{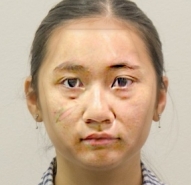} \\
        
        \rot{\rlap{~~~~~Victims}} & \includegraphics[width=30mm]{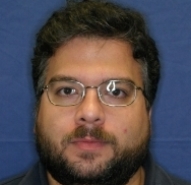} & \includegraphics[width=30mm]{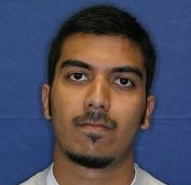} &
        \includegraphics[width=30mm]{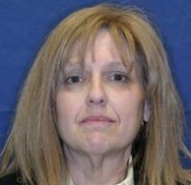} & \includegraphics[width=30mm]{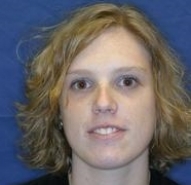} & \includegraphics[width=30mm]{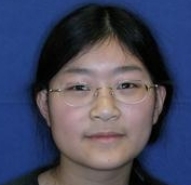} \\
        
        
         & a & b & c & d & e \\
    \end{tabular*}
    \caption{Examples of successful Break-in attacks: columns (a)-(b) with Attacker 1, (c)-(d) with Attacker 2, and (e) with Attacker 3. The FRS have misclassified the attackers as the subjects in the corresponding columns.}
    \label{fig:break_in}
\end{figure*}

\subsubsection{Training}
The face synthesizer needs to be trained with different variations of the Attacker's face images. Attackers use their individual face synthesizer models by training them with different variations of their face images (Figure \ref{fig:attackers}). For example, the synthesizer model for Attacker 1 is trained with four(4) modes each with multiple different labels. In total there are 48 training images for Attacker 1 which is the Cartesian product of all the modes and labels ($48 = 3\times4\times2\times2$):
\vspace{-2mm}
\begin{itemize}
    \itemsep-0.5em
    \item \textit{Facial hair}: clean, beard $+$ mustache, mustache only.
    \item \textit{Marks}: clean, forehead scar, cheek scar, mole.
    \item \textit{Expressions}: no expression, distorted cheek.
    \item \textit{Eyeglasses}: no eyeglasses, with eyeglasses.
\end{itemize}
These training images were preprocessed before applying the MMDA method to decompose. Preprocessing steps consist of locating facial landmarks, aligning the image to a reference face via the eye positions, normalizing the shape by applying a mask, and then to encode the warp face and landmarks coordinates in a vector $x$ (in Equation \ref{eqn:decompose}).
\vspace{-2mm}

\subsubsection{Decomposition}
The MMDA face synthesizer decomposes the training face images based on the different modes and labels within the modes. During the decomposition it learns two matrices $P$, and $V$. $P$ is the whitening PCA matrix. $V$ is an orthogonal matrix capturing semantic information of the modes and the shape of $V$ is $(n-1)\times(n-1)$ where $n$ is the number of training images. Here we show the following matrices only for the face synthesizer trained on Attacker 1's face images. Face synthesizer for other attackers follows similar calculations. Now $V$ is:
\begin{align}
    V &= [V^{face hair} \;  V^{mark} \;  V^{expression} \;  V^0]
\end{align}
The columns of the $V$ matrix are the bases for the facial hair ($V^{face hair}$), marks ($V^{mark}$), and expressions ($V^{expression}$). $V^0$ is the Residual Space. It captures the facial identity. Each face ($m$) can be decomposed into a vector $y$:
\begin{align}
    \label{eqn:decompose}
    y &= V^TP^Tx
    \\
    y^T &= [\underbrace{h_1 \, h_2}_\text{face hair} \;  \underbrace{m_1 \, m_2 \, m_3}_\text{marks} \;  \underbrace{e_1 \, e_2}_\text{expression} \;\underbrace{s^T}_\text{residual}]
    \\
    \label{eqn:p_s}
    p &= [h_1 \; h_2 \; m_1 \; m_2 \; m_3 \; e]
\end{align} 
During the attack, the parameter optimizer changes this vector to synthesize new face images apart from only the residual value. This is why the parameter vector $p$ for each Attacker does not have the residual space like the decomposed vectors $y^T$. Each of the attributes in the parameter vector $p$ (face hair, marks, etc.) need exactly one(1) less value than the number of labels within the attribute in the training images. The parameter vector $p$ has one more parameter $g$, which controls whether or not eyeglasses are synthesized. $g$ is separate from the MMDA face synthesizer. Eyeglasses are overlaid by using alpha-blending on the synthesized faces depending on the value of $g$.
\vspace{-4mm}

\subsubsection{Generating New Face} \label{goto:newface}
Given an altered parameter vector ${p'}$, the synthesis is achieved by: 
\begin{align}
\vspace{-3mm}
    x' &= P_rV{p'}
\end{align}
$V$ is the orthogonal matrix as described in the previous section and $P_r$ is learned during the decomposition. $P_r$ reverses the effects of $P$ in Equation \ref{eqn:decompose}. These vectors form a \textit{semantic basis} of the training space. Then the new face vector $x'$, which is a linear combination of the semantic faces, is reshaped and unwarped so that we are able to visualize the new face.

\begin{table}[]
\centering
\begin{tabular}{|>{}c|>{}c|}
\hline
$-0.50 < h_1, h_2   < 0.45$ & $-0.60 < m_1, m_2, m_3 < 0.50$ \\ \hline
\multicolumn{1}{|>{}c|}{\textbf{facial hair}} & \multicolumn{1}{>{}c|}{\textbf{marks}} \\ \hline \hline
$-0.40 < e < 0.40$ & $-0.20 < g < 0.20$ \\ \hline
\multicolumn{1}{|>{}c|}{\textbf{expression}} & \multicolumn{1}{>{}c|}{\textbf{eyeglasses}} \\ \hline
\end{tabular}
\caption{Bound constraints on the parameters in vector $p$ for Attacker 1. These help keep the synthesized face from being unrealistic.}
\label{tab:bounds}
\end{table}
The values in the altered parameter vector ${p}$ need to be bounded in order to prevent the newly synthesized image from being unrealistic. These  bounds on the parameters were selected from observing the learned whitening PCA matrix ($P$). Table \ref{tab:bounds}  shows the bounds on the parameters for Attacker 1. Similar bounds are set for the other attackers.

\section{Experiments and Results}

\subsection{Experiment setups} \label{goto:setup}
In the experiments we used the three FRS as mentioned in section \ref{goto:frs}. Gallery of 40 random face images were selected from the CMU Multi-PIE dataset\cite{gross2010multi} for majority of the experiments. For some experiments different galleries (from the same dataset) were selected based on the experiment.

\begin{figure}[h]
    \centering
    \includegraphics[width=\columnwidth]{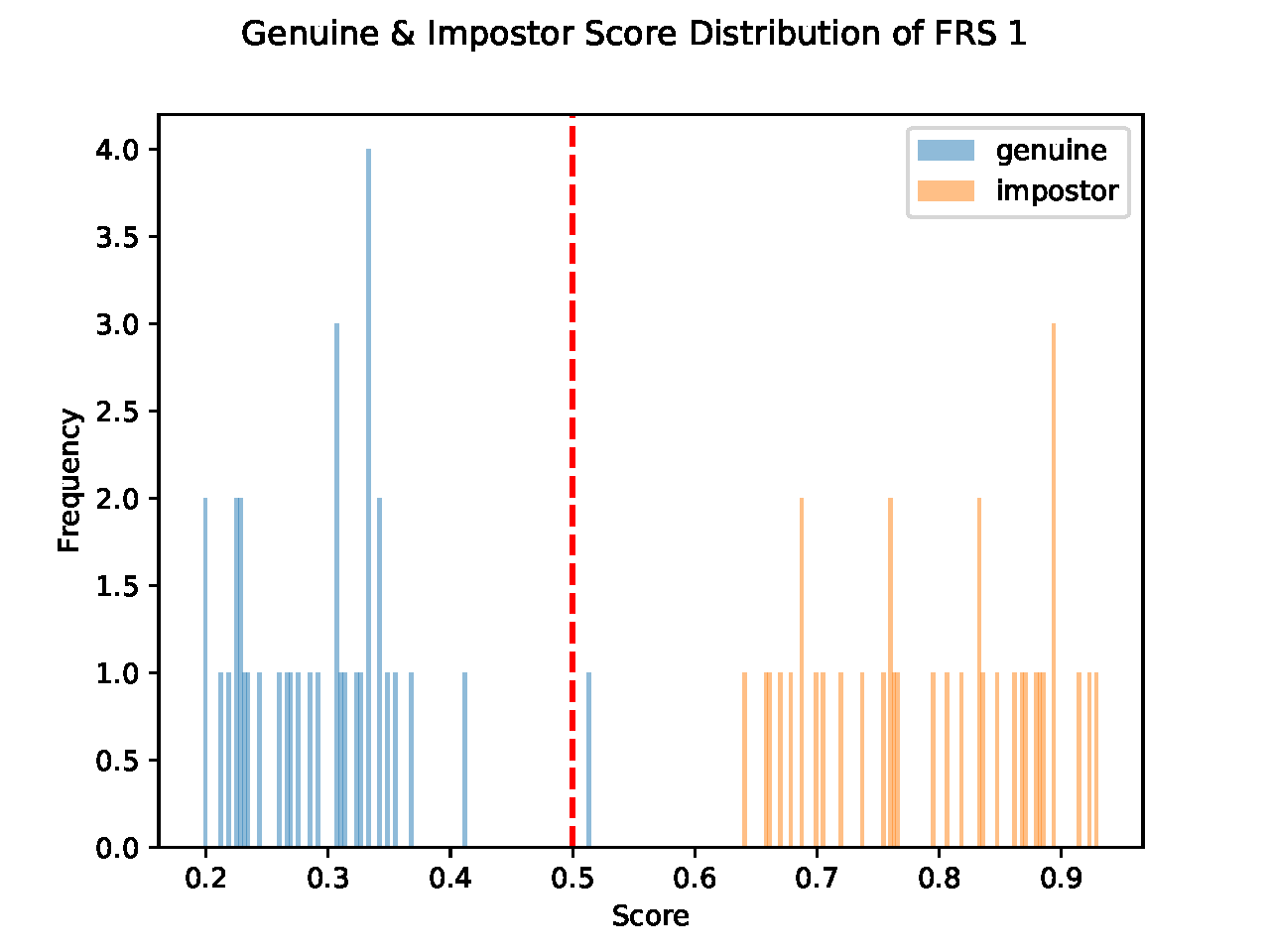}
    \caption{Genuine (match) and Impostor (non-match) Score Distribution for FRS 1: Python `\textit{face\_recognition}'). It justifies the selection of the threshold.}
    \label{fig:scoredist}
\end{figure}

Python \textit{face\_recognition} library and RESNET50 implementation returns euclidean and cosine \textit{distance} respectively. Lower \textit{distance} is considered better. The threshold $\theta$ in Equation \ref{eqn:optimization} for these two systems are $0.50$ and $0.45$ respectively. Hence, any attack with the final distance less than $\theta$ is considered a successful attack. Face++ API on the other hand returns \textit{confidence} (higher is better). The minimum \textit{confidence} for this system is $0.65$. Any attack with higher \textit{confidence} than this is considered successful. These thresholds values were taken as strict cutoff values from the official implementation instructions of the systems. We have also generated a `Genuine (match) and Impostor (non-match) Score Distribution' (see Figure \ref{fig:scoredist}) for one of the FRS (Python `\textit{face\_recognition}') with 35 subject in each group (genuine and impostor). This justifies the selection of the thresholds.

\subsection{Break-in}\label{goto:breakin}
In break-in attacks, the attacker aims to be accepted by the FRS as `anyone' from the \textit{gallery}. The initial parameters ($p_0$) for the synthesizer are randomly selected within the range of the bound constraints of Table \ref{tab:bounds}. Examples of the successful break-in attacks are presented in Figure \ref{fig:break_in}. The first row of images are of the attackers and the next row are the subjects the FRS identified the attacker as. The FRS misclassified the attack faces as the victims in the corresponding columns.

\subsection{Impersonation and Evasion}\label{goto:impersonation}
\begin{figure}
    \centering
    \begin{tabular*}{\textwidth}{ccccc}
        \rot{\rlap{~Attackers}} & \includegraphics[width=16mm]{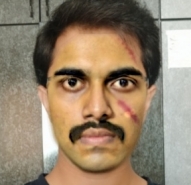} & \includegraphics[width=16mm]{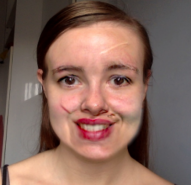} &
        \includegraphics[width=16mm]{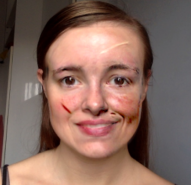} & \includegraphics[width=16mm]{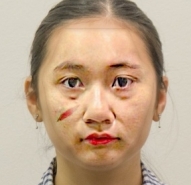} \\
        
        \rot{\rlap{~Victims}} & \includegraphics[width=16mm]{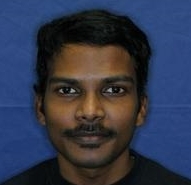} & \includegraphics[width=16mm]{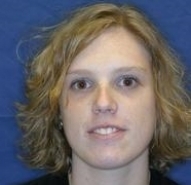} &
        \includegraphics[width=16mm]{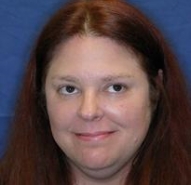} & \includegraphics[width=16mm]{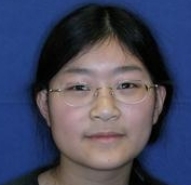} \\
        
        
         & a & b & c & d \\
    \end{tabular*}
    \caption{Examples of successful Impersonation attacks: column (a) with Attacker 1, (b)-(c) with Attacker 2, and (d) with Attacker 3. The attackers selected the victims prior to the attack and could successfully impersonate them.}
    \label{fig:impersonation}
\end{figure}

\textbf{Impersonation}.
In impersonation attacks the attacker tries to make the FRS classify him as `specifically someone' who he tries to impersonate. For Impersonation, the initial parameter vector $p_0$ was set to the parameters from decomposing Victim's face image with the face synthesizing method (Equations \ref{eqn:decompose} and \ref{eqn:p_s}). It is a reasonable starting point, as opposed to starting with a random $p_0$ like in break-in attacks. All three attackers could successfully impersonate multiple targeted victims. Figure \ref{fig:impersonation} shows some of the successful impersonation attempts. 

\begin{figure}[!h]
\centering
\begin{tabular}{c@{\hspace{1mm}}c:c@{\hspace{1mm}}c}
 \includegraphics[width=16mm]{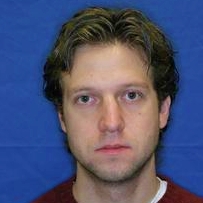} &  \includegraphics[width=16mm]{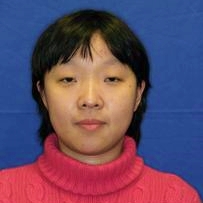} &
 \includegraphics[width=16mm]{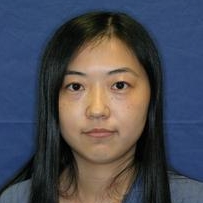} &
 \includegraphics[width=16mm]{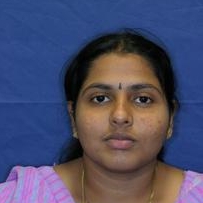} \\
 \multicolumn{2}{c}{(a)} & \multicolumn{2}{c}{(b)} \\
\end{tabular}
\caption{Victims for whom impersonation failed. We found that success in impersonation attacks depended on similarity of the Attacker and his Victims. Victims in (a), and (b) are the targets who could not be impersonated by Attacker 1, and Attacker 2 respectively.}
\label{fig:failedimpersonations}
\end{figure}

However, impersonation attacks succeed when the Attacker and the Victim have similar attributes (e.g. race, gender, facial hair etc.) between them. We also experimented with victims of different genders, and races for both the Attackers and reasonably failed to impersonate them i.e. the target FRS did not output the Victim's ID with a score below the threshold. Some of the victims whom we failed to impersonate are presented in Figure \ref{fig:failedimpersonations}.


\textbf{Evasion}\label{goto:evasion_res}. 
Evasion is more challenging task than the other two forms of attack. This is because the Attacker's own face is also in the \textit{gallery}. We have experimented with two different scenarios as discussed in section \ref{goto:evasions}. For the given \textit{gallery} and synthesizer, evasion from the \textit{gallery} was not successful in our experiments. In our experiments on evasion, the FRS were not fooled and they correctly identified the attacker's face.

\subsection{Attacking FRS with Real Face}
\begin{figure}[!htpb]
\centering
    \begin{tabular*}{\textwidth}{cc:ccc}
        \rot{\rlap{\footnotesize ~Attacker 1}} &
        \includegraphics[width=16mm]{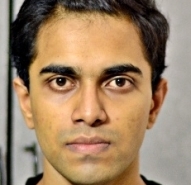} &
        \rot{\rlap{~Attackers}} &
        \includegraphics[width=16mm]{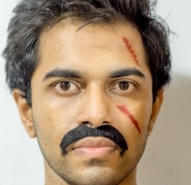} & \includegraphics[width=16mm]{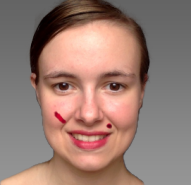} \\
        
        \rot{\rlap{\footnotesize Attacker 2}} &
        \includegraphics[width=16mm]{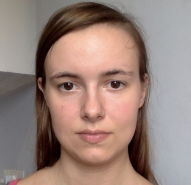} &
        \rot{\rlap{~Victims}} &
        \includegraphics[width=16mm]{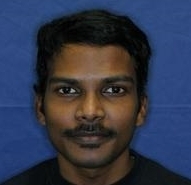} & \includegraphics[width=16mm]{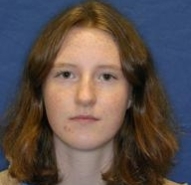} \\
        
        & Base faces &  & a & b \\
    \end{tabular*}
\caption{Faces in the left-most column are the base \textbf{real faces} of Attacker 1 and 2 which did not fool the FRS. In (a), Attacker 1 successfully impersonated the Victim by presenting his \textbf{actual face} (not an image) to the FRS by applying the scars and mustache informed by the synthesized image. In (b), Attacker 2 applied the makeup (synthesized in Figure \ref{fig:break_in}(c)) on her real face, and successfully broke into the FRS. The FRS misidentified Attacker 2 as the Victim shown.}
\label{fig:real}
\end{figure}

By replicating the makeups informed by the synthesized faces from some of the successful attacks, we presented the real face images to the FRS. Some of these results are shown in Figure \ref{fig:real}. We could successfully break-in and impersonate victims using real faces with makeups informed by the synthesized face images.

\subsection{Variations of Galleries}
\begin{figure}[h!]
 \centering
 \includegraphics[width=0.5\textwidth]{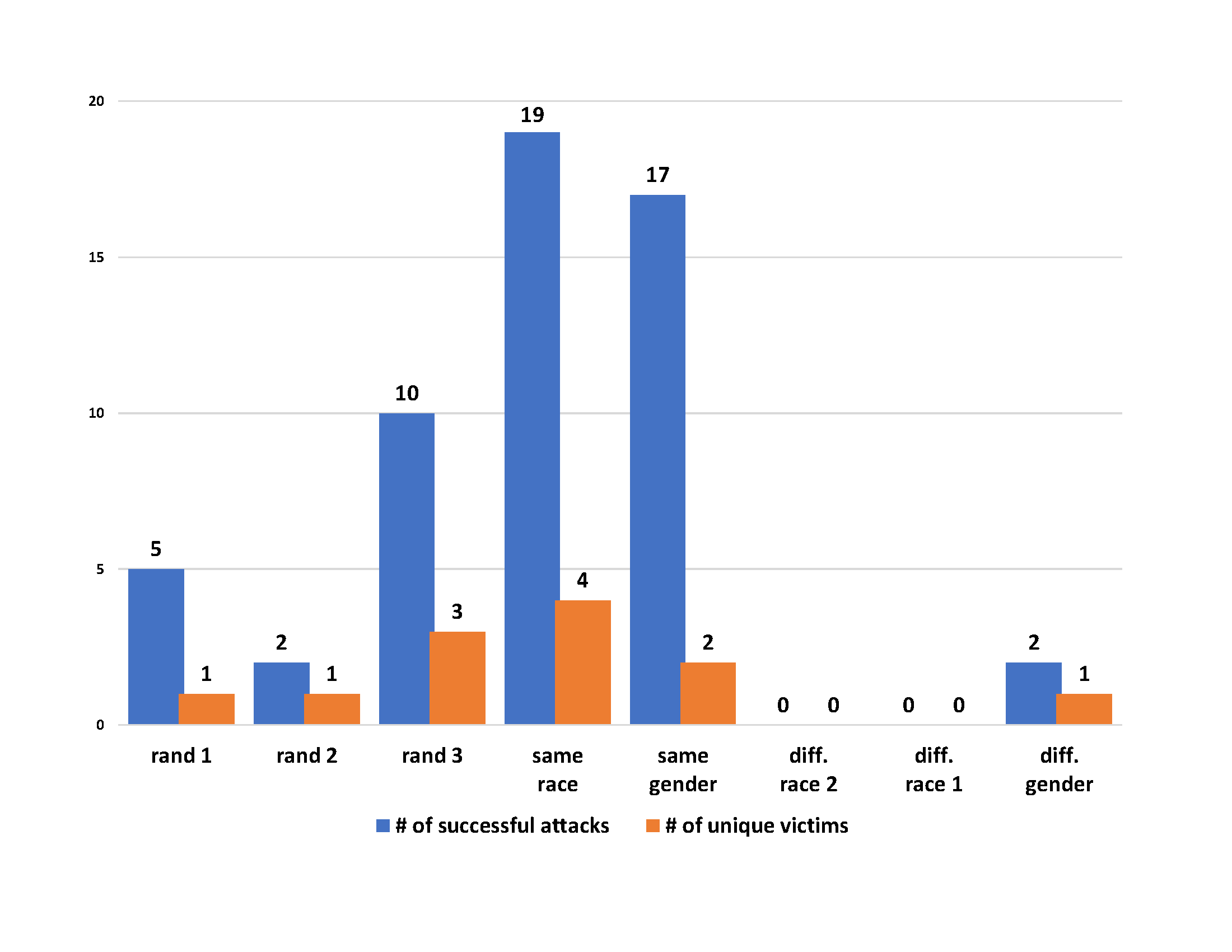}
 \caption{Impact of different galleries with varying difficulties: Number of successful attacks (out of 20 attempts) and number of unique victims for different variations of galleries. Break-in attacks are more successful when the \textit{gallery} has subject from same race, or gender as the attacker.}
 \label{fig:galleries}
\end{figure}

Following are some different combinations of galleries for break-in attacks: 
\begin{enumerate*}
    \item 3 random galleries of 50 individuals each.
    \item 2 galleries of 40 individuals from different races.
    \item 1 gallery of 50 individuals from different gender.
    \item 1 gallery of 40 individuals from same race.
    \item 1 gallery of 50 individuals from same gender.
\end{enumerate*}
In Figure \ref{fig:galleries} we plot number of successful attacks and unique victims from 20 attacks for different galleries as listed above. The figure shows that Break-in attacks succeed more often if the Attacker has the same gender and ethnicity as someone (anyone) in the gallery.

\subsection{Different Gallery Sizes}
\begin{figure}[h!]
 \centering
 \includegraphics[width=0.5\textwidth]{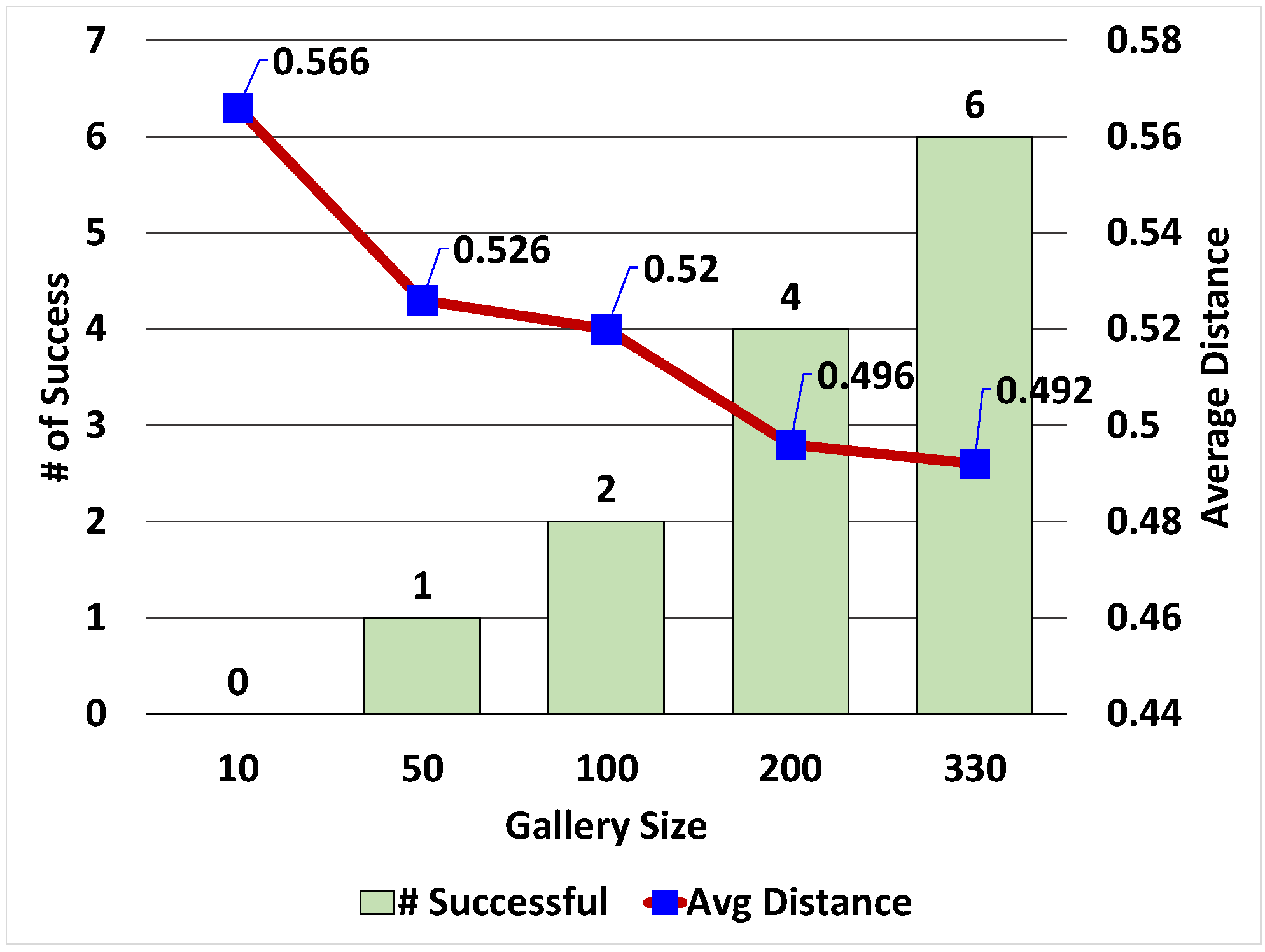}
 \caption{Impact of gallery size. For each \textit{gallery} size we plot the number of successful break-in attempts (in green bars) by Attacker 1, and average of minimum distances (lower is better) from 10 attempts. It is clear that larger \textit{gallery} size makes break-in easier.}
 \label{fig:gallerysize}
 \vspace{-3mm}
\end{figure}

We experimented the performance of proposed Attack Scheme with varying \textit{gallery} sizes. The aim is to find out if the size of \textit{gallery} has any impact. We made 10 attempts to break-in with Attacker 1 with \textit{gallery} sizes ranging from 10 to 330. The results of this experiments are given in Figure \ref{fig:gallerysize}. 

\subsection{Variations in Face Synthesizer}
\textbf{Without Bound Constraints}.
When attack-faces are synthesized without the bound constraints (e.g. Table \ref{tab:bounds} - for Attacker 1) on the synthesizing parameters, the face images sometimes turn out to be unrealistic. This is due to the parameter optimizer pushing the parameters too far. However, synthesizing faces without these bounds generates more successful attacks. This can be useful in a scenario when attacks on FRS can be done with an image rather than an actual face (e.g. online upload). Although not realizable, these images fool the FRS successfully.

\textbf{More Training Images}. 
Training the face synthesizer with more training images (e.g. face with different types of scars, moles, makeups, etc.) creates more space for the synthesizer to generate a new face. Moreover, higher number of successful attacks are generated. However, having more facial attributes would mean there are more things to take care of during the attack with real faces. The number of combinations of the training images (for face synthesizer) is a trade-off between getting more successful in the attacks, and synthesizing face which are easier to realize.

\subsection{Scaling to Multiple FRS}
Experiments were done with the three FRS listed in Section \ref{goto:frs}. Multiple experiments with all three attackers show that these FRS are prone to misclassifying attack faces for the three attackers. Hence, it is evident that our attack scheme scales to multiple FRS with multiple attackers.

\section{Discussion and Conclusion}
From the experiments we counted that approximately $21.8\%$ of the attacks (Break-in and Impersonation) were successful for the three attackers combined. Hence, it is clear that face recognition systems are vulnerable to our attack scheme. We showed that real perturbations of the attackers' face succeed in break-in and impersonation attacks. However, evasion was not successful as the other two types of attacks. Moreover, impersonation attacks depend on the similarity between the Attacker and the Victim. Getting a good initial parameter vector also plays an important role in finding an effective final image. We noticed that increasing gallery size improves the chances of success. Also, a gallery with faces of similar race, gender as the attacker increases the chances of a successful attack. Finally, we have seen that attackers were successful in attacking with real face which had been realized from a synthesized image. This shows that our attack scheme successfully fooled the FRS in a black-box settings with real faces.

Our work makes a clarion call for urgent research to address these vulnerabilities in FRS. We hope other researchers will take up this challenge.

{\small
\bibliographystyle{ieee}
\bibliography{reference}
}

\end{document}